\documentclass{article}
\usepackage[utf8]{inputenc}

\usepackage{PRIMEarxiv}

\usepackage[utf8]{inputenc} 
\usepackage[T1]{fontenc}    
\usepackage{hyperref}       
\usepackage{url}            
\usepackage{booktabs}       
\usepackage{amsfonts}       
\usepackage{nicefrac}       
\usepackage{microtype}      
\usepackage{lipsum}
\usepackage{fancyhdr}       
\usepackage{graphicx}       
\graphicspath{{media/}}     
\usepackage{amsmath}
\title{
Concept-Driven Deep Learning for Enhanced Protein-Specific Molecular Generation
}

\author{
  Taojie Kuang \\
  Peng Cheng Laboratory\\
  South China University of Technology\\
  \texttt{kuangtj@pcl.ac.cn} \\
  \And
  Qianli Ma \\
  South China University of Technology\\
  \texttt{qianlima@scut.edu.cn} \\
  \And
  Athanasios V. Vasilakos \\
  University of Agder\\
  \texttt{Thanos.vasilakos@uia.no}\\
  \And
  Yu Wang\\
  Peng Cheng Laboratory\\
  \texttt{wangy20@pcl.ac.cn}\\
  \And
  Qiang (Shawn) Cheng\thanks{Corresponding author} \\
  U Kentucky\\
  \texttt{Qiang.Cheng@uky.edu} \\
  \And
  Zhixiang Ren\thanks{Corresponding author} \\
  Peng Cheng Laboratory\\
  \texttt{jason.zhixiang.ren@outlook.com} \\
}

\begin{document}
\maketitle

\begin{abstract} 
In recent years, deep learning techniques have made significant strides in molecular generation for specific targets, driving advancements in drug discovery. 
However, existing molecular generation methods present significant limitations: those operating at the atomic level often lack synthetic feasibility, drug-likeness, and interpretability, while fragment-based approaches frequently overlook comprehensive factors that influence protein–molecule interactions. 
To address these challenges, we propose a novel fragment-based molecular generation framework tailored for specific proteins. 
Our method begins by constructing a protein subpocket and molecular arm concept-based neural network, which systematically integrates interaction force information and geometric complementarity to sample molecular arms for specific protein subpockets. 
Subsequently, we introduce a diffusion model to generate molecular backbones that connect these arms, ensuring structural integrity and chemical diversity. 
Our approach significantly improves synthetic feasibility and binding affinity, with a 4\% increase in drug-likeness and a 6\% improvement in synthetic feasibility.
Furthermore, by integrating explicit interaction data through a concept-based model, our framework enhances interpretability, offering valuable insights into the molecular design process.

\end{abstract}


\section{Introduction}
\label{sec:intro}

Structure-based molecular generation uses 3D protein structures to design ligands with high binding affinities, a key approach in modern drug discovery. Traditional methods, like high-throughput screening, are costly, slow, and constrained by predefined libraries. Deep learning addresses these limitations by automating design, efficiently exploring chemical spaces, and improving molecular property predictions, enabling faster and more effective drug discovery\cite{zeng2022deep, isert2023structure}.

Autoregressive models were among the first deep learning approaches applied to structure-based molecular generation, such as GraphBP\cite{liu2022generating} and Pocket2Mol\cite{peng2022pocket2mol}, which sequentially predict atom types and positions to construct ligands based on protein binding site information. 
With the success of diffusion models in computer vision, they have been adapted to SBDD to address the limitations of autoregressive methods. 
Diffusion models, such as DiffBP\cite{lin2025diffbp}, TargetDiff\cite{guan20233d}, and DiffSBDD\cite{schneuing2024structure}, employ E(3)-equivariant diffusion frameworks to model protein and molecular information, generating molecules through iterative denoising processes that ensure chemical validity and binding affinity. 
Additionally, MOOD\cite{lee2023exploring} expands the exploration of chemical space by integrating out-of-distribution controls, enabling the discovery of novel and diverse molecules. 
Despite these advancements, atom-based methods often produce impractical molecules due to overlooked synthetic feasibility and drug-likeness, limiting their applicability in real-world scenarios.

Fragment-based molecular generation constructs drug molecules from small, biologically meaningful fragments, improving efficiency, synthetic feasibility, and drug-likeness. 
Methods like MiCaM\cite{geng2023novo} and DECOMPDIFF\cite{guan2024decompdiff} leverage motif mining and diffusion models to assemble arms and scaffolds, enhancing binding affinity and stability. 
Other approaches, such as D3FG\cite{lin2024functional} and HierDiff\cite{qiang2023coarse} use functional fragments or coarse geometries to generate, employ functional fragments and hierarchical modeling for protein-specific molecular generation. 
Reinforcement learning frameworks like FREED\cite{yang2021hit} further optimize fragment-based leads, while tools such as DeepFrag\cite{green2021deepfrag} and GEAM\cite{lee2023drug} refine molecules through fragment reconstruction and goal-oriented extractions. However, many methods neglect critical protein-molecule interactions, such as geometric complementarity and interaction forces, limiting their ability to fully capture binding dynamics.

\begin{figure}[t]
    \centering
    \includegraphics[width=1.0\textwidth]{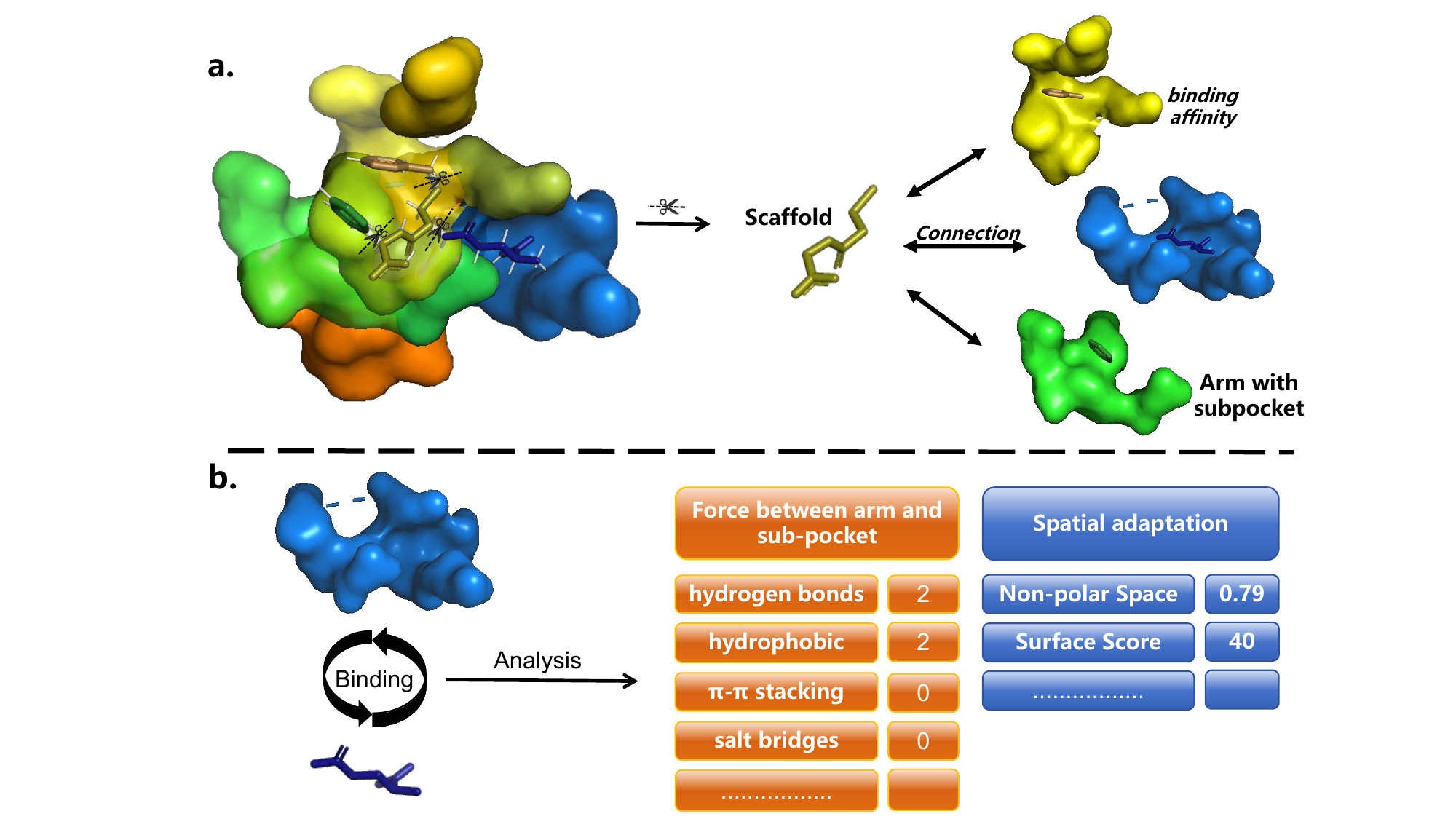}  
    \caption{\textbf{The influence factors on the affinity between the protein sub-pocket and the ligand arm.} a. The binding affinity between the protein pocket and the ligand can be divided into two main parts: the binding between the ligand arm and the protein sub-pocket, and the overall binding between the molecule and the protein. b. The affinity between the ligand arm and the protein sub-pocket is influenced by two main factors: spatial adaptation and non-covalent bonds. Spatial adaptation refers to how well the ligand arm fits into the protein sub-pocket, while non-covalent bonds include interactions like hydrogen bonds, hydrophobic forces, $\pi-\pi$ stacking, and salt bridges. These factors collectively determine the overall binding strength.}
    \label{fig:intro}
\end{figure}

As the Figure \ref{fig:intro} Ligands can be conceptually decomposed into scaffolds and arms, where arms interact with the protein target to achieve high binding affinity, and scaffolds position the arms into the desired binding regions\cite{guan2024decompdiff}. The arms correspond to subpockets within the protein binding site, and selecting arms that fit a subpocket depends on two key factors: the interaction forces between atoms and the geometric complementarity between the arm’s surface and the subpocket. We seek to incorporate these critical factors into a unified model for improved ligand design.

We begin by extracting a large set of molecular arms and protein sub-pocket pairs from protein-ligand molecule databases and chemical databases. Using various software tools\cite{salentin2015plip}, we analyze these pairs to generate detailed interaction force information and assess the geometric complementarity between the molecular arms and the protein sub-pockets. This data forms the foundation for constructing a neural network framework that leverages the concepts of protein subpockets and molecular arms, systematically integrating these factors to select the optimal molecular arms for specific protein subpockets. 
Subsequently, we employ a diffusion model to generate molecular backbones that connect these arms, ensuring structural integrity and chemical diversity. 
This approach demonstrates significant advantages in synthetic feasibility and binding affinity compared to traditional methods.
Furthermore, the concept-based model enhances interpretability in arm selection, clarifying why specific arms are chosen for subpockets.
The followings are our highlight:
\begin{itemize}
\item[$\bullet$] We incorporate a protein-subpocket and molecular-arm interaction-aware knowledge graph to guide molecular generation based on biologically meaningful interactions.
\item[$\bullet$] We use concept-based models to sample molecular arms aligned with interpretable and functionally relevant features of the target protein.
\item[$\bullet$] Our approach demonstrates significant improvements in synthetic feasibility, binding affinity, and interpretability compared to existing methods. Specifically, the molecules generated by our method show a 4\% increase in drug-likeness and a 6\% improvement in synthetic feasibility. 
\end{itemize}

\section{Related Work}
\label{sec:Rel}
\subsection{Structure-Based Drug Design(SBDD)}

SBDD leverages 3D structural information of proteins to design molecules with strong binding affinities, serving as a cornerstone of modern drug discovery. Recent advancements in deep learning have greatly enhanced this field, offering more precise and efficient methods for molecule generation.
Autoregressive Methods like GraphBP\cite{liu2022generating} and Pocket2Mol\cite{peng2022pocket2mol} employ 3D graph neural networks and E(3)-equivariant frameworks to predict atom types, positions, and bonding, aligning molecular fragments with protein pockets. G-SphereNet\cite{luo2022autoregressive} further refines molecular geometry by predicting distances, angles, and torsions while maintaining rotational invariance. MolCode\cite{zhang2023equivariant} integrates autoregression with flow-based modeling to co-design molecular graphs and 3D structures, while DeepICL\cite{zhung20243d} incorporates interaction-aware contexts to enhance atom placement. However, these approaches often face challenges like error accumulation and reliance on generation order.
Variational Autoencoders (VAEs) enable efficient molecular sampling, with GF-VAE\cite{ma2021gf} combining flow-based decoders for one-shot graph generation and Atomic Grid VAEs reconstructing ligand conformations from density grids. 
Diffusion Models have emerged as transformative tools in SBDD. TargetDiff\cite{guan20233d}, DiffSBDD\cite{schneuing2024structure}, and DiffBP\cite{lin2025diffbp} utilize SE(3)-equivariant diffusion to jointly generate molecular coordinates and atom types, ensuring chemical validity and binding affinity. 
DiffPROTACs\cite{li2024diffprotacs} uses O(3)-equivariant graph Transformers to design PROTACs, while Diff-AMP\cite{wang2024diff} focuses on antimicrobial peptide generation by optimizing stability, diversity, and bioactivity. 
InterDiff\cite{wu2024guided} incorporates ligand-protein interaction prompts, tailoring molecules for specific protein environments, and MOOD\cite{lee2023exploring} explores out-of-distribution chemical spaces while optimizing drug-likeness and binding affinity through property-guided diffusion.
Despite these advancements, atom-based methods often produce impractical molecules due to overlooked synthetic feasibility and drug-likeness. Additionally, their limited interpretability poses challenges for real-world applications.

\subsection{Fragment-Based Drug Design(FBDD)}

FBDD posits that a biologically active drug molecule results from the combined bioactivity of multiple small fragments. Its key advantages lie in efficiency and precision, as it simplifies drug design by focusing on meaningful fragments, accelerating chemical space exploration, and ensuring the generation of realistic and pharmacologically viable molecules. 
Recently, numerous studies\cite{huang20223dlinker, imrie2021deep, igashov2024equivariant} have adopted this approach to generate molecules effectively.
Methods such as MiCaM\cite{geng2023novo}, which mines frequent motifs from datasets to construct realistic structures, and DECOMPDIFF\cite{guan2024decompdiff}, which employs diffusion models to decompose ligands into functional arms and scaffolds, ensure high binding affinity and stability. DeepFrag\cite{green2021deepfrag} optimizes ligands by reconstructing missing fragments, while GEAM\cite{lee2023drug} dynamically extracts goal-specific fragments to refine drug candidates.
Advanced diffusion-based methods like D3FG\cite{lin2024functional} and HierDiff\cite{qiang2023coarse} use functional fragments or coarse geometries to generate realistic 3D molecular structures optimized for protein binding. Similarly, FREED\cite{yang2021hit} integrates reinforcement learning with fragment-based generation for pharmacochemical validity, and MoLeR\cite{maziarz2021learning} extends scaffolds with structural motifs for efficient scaffold-based tasks. PS-VAE\cite{kong2022molecule} and FLAG\cite{zhang2023molecule} leverage principal subgraphs and cheminformatics tools, respectively, to refine molecular geometries and improve structural quality.
While these methods advance the field, they often neglect key protein–molecule interactions. Our method fully integrates interaction forces and geometric complementarity, boosting binding affinity and interpretability.

\section{Method}

\subsection{Overview}

\begin{figure}[t]
    \centering
    \includegraphics[width=0.82\textwidth]{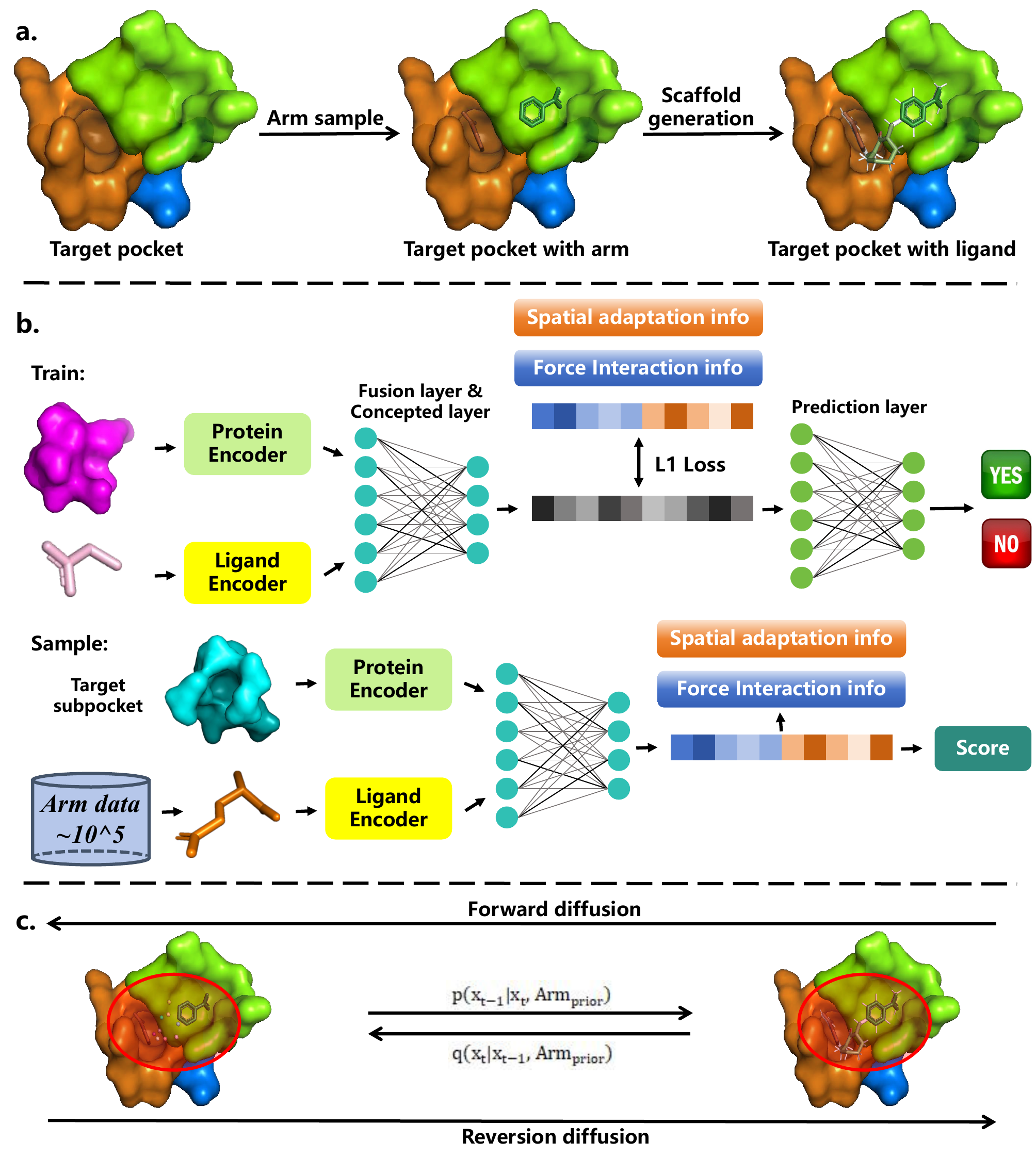}  
    \caption{\textbf{The overview of our method. }
    \textbf{a. Two-stage process for ligand molecular generation.} First, ligand arms are sampled based on their compatibility with the target protein subpocket. Then, the scaffold is generated by linking the chosen arms, forming a complete molecular structure that fits the protein pocket. 
    \textbf{b. Ligand arm sampling.} The first stage involves arm sampling, where a concept-based neural network identifies molecular arms that fit the subpockets of a target protein based on their interaction forces and geometric complementarity. Spatial adaptation and force interaction information are used to guide the selection of the most suitable arms for the protein subpocket.
    \textbf{c. Scaffold generation.} The second stage uses a diffusion model to generate molecular backbones that connect the selected arms, ensuring the ligands are structurally intact, diverse, and compatible with the protein binding pocket. The forward and reverse diffusion processes, combined with protein and ligand encoders, refine the scaffold structure, incorporating atom and bond types of arm as prior knowledge.}
    \label{fig:method}
\end{figure}

Our proposed framework for protein-specific molecular generation addresses key challenges in drug discovery by integrating protein subpockets and molecular arm concepts into a two-stage generative pipeline. As the Figure \ref{fig:method} the first stage involves arm sampling, where a concept-based neural network identifies molecular arms that fit the subpockets of a target protein based on their interaction forces and geometric complementarity. The second stage focuses on scaffold generation, where a diffusion model generates molecular backbones that connect these arms, ensuring the resulting ligands are structurally intact, chemically diverse, and compatible with the protein binding pocket.


In the arm sampling stage, we first extract and process a large dataset of protein-ligand molecular pairs from protein-ligand databases. From these datasets, we obtain protein sub-pocket and molecular arm pairs, which are analyzed using various software tools to generate key concept information. Subpockets are defined as localized regions within the protein's binding pocket that interact with specific parts of the ligand (arms). By encoding the 3D structure and chemical properties of the subpocket and arms into latent representations, our model predicts an arm-subpocket compatibility score. This score guides the selection of arms from a molecular fragment library, ensuring high binding affinity and proper geometric fit for each subpocket. This stage not only streamlines the arm selection process but also provides interpretability, as the model explicitly highlights the key features influencing arm-subpocket compatibility.

The scaffold generation stage uses a diffusion model to generate molecular backbones that connect the selected arms. Starting from a random noise distribution, the diffusion model progressively refines the molecular structure through iterative denoising steps. The model incorporates an E(3)-equivariant framework to ensure that the generated scaffolds are invariant to rotation and translation, critical for accurate 3D ligand generation. This stage ensures that the scaffold maintains structural integrity by forming stable chemical bonds with the arms, chemical diversity by exploring a wide range of possible scaffolds, and binding compatibility by fitting well within the overall protein binding pocket.

Together, these two stages enable our framework to produce protein-specific ligands with enhanced binding affinity, synthetic feasibility, and interpretability, addressing limitations in traditional and contemporary molecular generation methods.

\subsection{Concept based model for arm sampling}
Protein-ligand binding can be effectively understood by breaking it down into interactions between protein subpockets and molecular arms. In this framework, molecular ligands are divided into two components: molecular arms, which interact directly with protein subpockets to achieve binding, and molecular scaffolds, which serve as structural backbones connecting the arms. To optimize molecular arm selection, we define a set of concepts that represent critical factors influencing the interaction between molecular arms and protein subpockets. These concepts are grouped into two categories: spatial factors and interaction forces.

\subsubsubsection{\textbf{Subpocket-Arm Data Processing}}
Our data comes from the PDBbind2020\cite{liu2017forging} and Crossdocked\cite{francoeur2020three} datasets, from which we select high-quality docking poses (RMSD < 1 Å) and ensure protein diversity (sequence identity < 30\%). Using the Protein-Ligand Interaction Profiler (PLIP)\cite{salentin2015plip}, we generate the protein sub-pockets based on their interaction regions. Afterward, molecular fragments are assigned to these subpockets based on their spatial location and linked to create ligand arms. These molecular fragments are generated according to the breaking of retrosynthetically interesting chemical substructure (BRICS)\cite{degen2008art} fragmentation rules. This results in a subpocket-arm dataset consisting of approximately 400K arm-subpocket pairs, which serve as the foundation for training our concept-based model. Importantly, these molecules do not include the novel proteins used for testing during the molecular generation phase.

\subsubsubsection{\textbf{Concept layer and key interaction factors}} 
The binding between molecular arms and protein subpockets can be understood through two key concepts: Spatial factors and Interaction forces. Spatial factors describe the physical compatibility of the molecular arm with the protein subpocket, while Interaction forces reflect the strength and types of interactions between them.
Spatial factors include non-polar space occupation, which measures the proportion of the subpocket’s non-polar space occupied by the molecular arm, and surface complementarity, which evaluates how well the surface of the molecular arm matches the geometry of the subpocket, ensuring minimal steric clashes and optimal binding orientation.
Interaction forces focus on the various interactions that occur between the molecular arm and the protein subpocket. The majority of these interactions are non-covalent bonds, such as hydrogen bonds, hydrophobic interactions, $\pi$-$\pi$ stacking, salt bridges, water bridges, halogen bonds, and $\pi$-cation interactions. These non-covalent interactions are crucial for molecular binding, as they are typically responsible for the stability and specificity of the ligand-protein interactions. The PLIP\cite{salentin2015plip} is used to calculate these interaction forces, providing a detailed measure of binding strength based on the structural and chemical characteristics of the interacting molecules.

\subsubsubsection{\textbf{Embedding for the molecular arm and protein subpocket}} 
To extract the relevant features for these key factors, we process data from a protein-ligand binding database. Molecular arms are encoded using a SchNet\cite{schutt2018schnet} encoder, which captures atomic and geometric properties of the arms, while protein subpockets are encoded using a Continuous-Discrete Convolutio\cite{fan2022continuous} encoder to model their structural and chemical characteristics. These encodings are then integrated into a concept-based model trained to predict the key factors influencing arm-subpocket binding. 
\[
\mathbf{E}_{arm} = \text{SchNet}(A_{arm})
\]
\[
\mathbf{E}_{pocket} = \text{CDC}(P_{pocket})
\]
\[
\mathbf{C}_{spatial} = f_{spatial}(\mathbf{E}_{arm}, \mathbf{E}_{pocket})
\]
\[
\mathbf{C}_{interaction} = f_{interaction}(\mathbf{E}_{arm}, \mathbf{E}_{pocket})
\]
where \( \mathbf{E}_{arm} \) and \( \mathbf{E}_{pocket} \) represented the embedding of arm and pocket. \( f_{spatial} \) is a function that predicts spatial factors from the arm and subpocket embeddings, and \( f_{interaction} \) predicts interaction forces between the arm and the subpocket based on the same embeddings.

\subsubsubsection{\textbf{Loss function}} 
For the spatial factors (non-polar space occupation and surface complementarity), we use L2 loss function to compare the predicted values with the actual values, and use the Poisson negative log-likelihood loss for the interaction forces.
The loss function are as followings:
\[
\mathcal{L}_{spatial} = \| \mathbf{C}_{spatial} - \mathbf{C}_{spatial, true} \|_2^2
\]
\[
\mathcal{L}_{interaction, i} = -\left[ y_i \log(\hat{\lambda}_i) - \hat{\lambda}_i \right]
\]
\[
\mathcal{L}_{total} = \lambda_{spatial} \mathcal{L}_{spatial} + \sum_i \lambda_{interaction, i} \mathcal{L}_{interaction, i}
\]
Where 
\( \mathbf{C}_{spatial} \) is the predicted spatial factor from the concept layer.
\( \mathbf{C}_{spatial, true} \) is the ground truth value for the spatial factor.
\( y_i \) is the actual number of interactions (e.g., number of hydrogen bonds).
\( \hat{\lambda}_i \) is the predicted rate or expected number of interactions for interaction type \( i \).
\( \lambda_{spatial} \) and \( \lambda_{interaction, i} \) are the weights assigned to the spatial and interaction loss terms, respectively.
The sum over \( i \) accounts for different interaction types, each with its own loss.
This model provides insights into the compatibility of molecular arms with specific subpockets.

\subsubsubsection{\textbf{Sampling and selection}} 
After training, given a new protein subpocket, the molecular arm library is filtered based on the subpocket's atomic capacity, and for each candidate arm, we calculate the predicted interaction factors, including spatial factors and interaction forces, using the trained model. A compatibility score is then computed by combining these factors, and the arm with the highest score is selected. The following are the score function:
\[
S_{compat} = \lambda_{spatial} \cdot \mathbf{C}_{spatial} + \sum_i \lambda_{interaction, i} \cdot \mathbf{C}_{interaction, i}
\]
where \( S_{compat} \) is the score computed by combining these factors, \( \mathbf{C}_{spatial} \) represents the predicted spatial factors (such as non-polar space occupation and surface complementarity), \( \mathbf{C}_{interaction, i} \) represents the predicted interaction forces for each type \( i \), \( \lambda_{spatial} \) and \( \lambda_{interaction, i} \) are the weights of score function, balancing the contributions of spatial and interaction factors to the compatibility score.
This ensures that the chosen molecular arms are geometrically compatible with the subpocket and optimized for interaction strength, making the design process both efficient and interpretable.

By systematically considering spatial compatibility and interaction forces, this concept-based approach ensures that the selected molecular arms are not only physically and chemically compatible with the protein subpocket but also optimized for binding affinity. Additionally, the concept-driven nature of the model enhances interpretability, providing clear insights into the rationale behind arm selection for specific subpockets.

\subsection{Diffusion model for scaffold generation}
Once the molecular arms are selected, the next step is to generate a scaffold that connects these arms into a complete ligand. This process is carried out using a diffusion model, which frames scaffold generation as a denoising problem. The model relies on prior knowledge obtained from the concept-based model, including the atomic properties of the sampled molecular arms, the bond attributes, and the positions of the reference arm centers, along with the structural information of the target protein. This combination of prior knowledge helps guide the diffusion process, ensuring that the generated scaffold is both chemically valid and geometrically compatible with the target protein. The diffusion model works by starting with a random noise vector \( \mathbf{z}_T \), which represents a fully noisy scaffold, and then iteratively refining the scaffold towards a clean structure \( \mathbf{z}_0 \) through multiple denoising steps. Each step progressively reduces the noise, guided by the prior knowledge and the task-specific information (such as arm-subpocket compatibility). This iterative refinement process is modeled as a stochastic differential equation and can be split into two phases: the forward process (where noise is added) and the reverse process (where noise is removed).

\subsubsubsection{\textbf{Forward Process}} 
In the forward process, noise is added to the scaffold by progressively corrupting the atomic positions and bond properties over \( T \) timesteps. The forward process for each timestep \( t \) is described by the conditional distribution \( q(\mathbf{z}_t | \mathbf{z}_{t-1}) \), which adds Gaussian noise to the scaffold at each timestep:

\[
q(\mathbf{z}_t | \mathbf{z}_{t-1}) = \mathcal{N}(\mathbf{z}_t; \mathbf{z}_{t-1}, \beta_t \mathbf{I})
\]

Where \( \mathcal{N}(\cdot; \mu, \sigma^2) \) is the Gaussian distribution with mean \( \mu \) and variance \( \sigma^2 \), \( \mathbf{z}_t \) represents the molecular scaffold at timestep \( t \), \( \beta_t \) is the noise schedule that controls the noise level at timestep \( t \).

For each timestep, noise is added to the atomic positions of the scaffold \( \mathbf{z}_t \) based on the atom’s location relative to the center of the arm's position. Additionally, the attributes of the bonds between atoms are corrupted by discrete noise, reflecting the uncertainty in the bond structure at each stage. The noise is added according to the following procedure:
Atomic position noise: For each atom in the scaffold, Gaussian noise is added to the atomic positions based on its location relative to the center of its corresponding molecular arm.
Bond attribute noise: The properties of the bonds between atoms (such as bond length, angle, etc.) are corrupted with discrete noise.

\subsubsubsection{\textbf{Reverse Process}} 
In the reverse process, the model learns to iteratively remove noise from the scaffold to recover the clean structure \( \mathbf{z}_0 \). The denoising process is conditioned on prior knowledge, including the atomic properties, bond attributes, and the target protein structure. This allows the model to use the reference arm positions and the protein-ligand interaction information to guide the denoising process. The reverse process is modeled by the following conditional distribution:

\[
p_\theta(\mathbf{z}_{t-1} | \mathbf{z}_t, \mathbf{C}_{prior}) = \mathcal{N}(\mathbf{z}_{t-1}; \mu_\theta(\mathbf{z}_t, t, \mathbf{C}_{prior}), \sigma_t^2 \mathbf{I})
\]

Where \( \mathbf{C}_{prior} \) represents the prior knowledge, which includes the atomic properties, bond attributes, reference arm positions, and structural information of the target protein, \( \mu_\theta(\mathbf{z}_t, t, \mathbf{C}_{prior}) \) is the mean predicted by the neural network at timestep \( t \), conditioned on the noisy scaffold \( \mathbf{z}_t \) and the prior knowledge \( \mathbf{C}_{prior} \), \( \sigma_t^2 \) is the variance of the noise at timestep \( t \). During the reverse process, the model learns to progressively reduce the noise, using prior knowledge to guide the denoising and ensure that the scaffold is compatible with the binding pocket of the target protein.

\subsubsubsection{\textbf{Loss Function}} 
The training of the diffusion model is done by maximizing the likelihood of generating a clean scaffold \( \mathbf{z}_0 \) from a noisy scaffold \( \mathbf{z}_T \), conditioned on the prior knowledge. This is achieved by minimizing the variational lower bound (VLB), which is equivalent to the following loss function:

\[
\mathcal{L}_{diff} = \mathbb{E}_{q} \left[ \| \mathbf{z}_0 - p_\theta(\mathbf{z}_0 | \mathbf{z}_T, \mathbf{C}_{prior}) \|_2^2 \right]
\]

Where \( p_\theta(\mathbf{z}_0 | \mathbf{z}_T, \mathbf{C}_{prior}) \) is the model’s prediction for the clean scaffold \( \mathbf{z}_0 \), conditioned on the noisy scaffold \( \mathbf{z}_T \) and the prior knowledge \( \mathbf{C}_{prior} \), The expectation \( \mathbb{E}_{q} \) is taken over the noisy samples from the forward process. This loss function encourages the model to generate scaffolds that are chemically valid, structurally sound, and compatible with the protein-ligand binding pocket, while also ensuring diversity in the generated scaffolds.





\section{Experiments and Result}

\subsection{Experimental Setup}
\subsubsubsection{\textbf{Dataset}}
Following previous works such as Peng et al.\cite{peng2022pocket2mol} and Guan et al.\cite{guan20233d}, we utilize the CrossDocked2020\cite{francoeur2020three} dataset, which contains 22.5 million docked protein-ligand complexes. 
To ensure data quality and relevance, we filter the dataset by selecting only complexes with root mean square deviation (RMSD) less than 1 Å between the docked pose and the ground truth, as well as protein sequences with less than 30\% sequence identity. 
This refinement results in a subset of approximately 100,000 high-quality protein-ligand pairs for training. 
For evaluation, we use 100 novel protein-ligand complexes as test references.
\subsubsubsection{\textbf{Baseline}}
We compare our proposed method with several state-of-the-art molecular generation approaches:
AR\cite{luo20213d} are GNN-based methods that generate 3D molecules atom by atom in an autoregressive manner.
GraphBP\cite{liu2022generating}: A graph neural network-based autoregressive approach that sequentially constructs ligands by predicting atom types and positions.
TargetDiff\cite{guan20233d}: A diffusion model leveraging E(3)-equivariant frameworks to model protein-ligand interactions and generate molecules with high binding affinities.
DecompDiff\cite{guan2024decompdiff}: A fragment-based diffusion model that decomposes ligands into functional arms and scaffolds to enhance structural diversity and binding stability.
These baselines represent a diverse range of generative techniques, including autoregressive, diffusion, GAN, and fragment-based approaches, allowing for a comprehensive comparison of our method against existing strategies.
\subsubsubsection{\textbf{Evaluation}}
We evaluate the generated molecules from two perspectives: target binding affinity and molecular properties. For binding affinity, we use AutoDock Vina\cite{eberhardt2021autodock} to estimate binding scores, following the protocols established by Luo et al.\cite{luo20213d} and Ragoza et al.\cite{ragoza2022generating}. Key metrics include the Vina Score, which directly estimates the binding affinity of the generated 3D molecular structures, and Vina Min, which calculates binding scores after local structural minimization to account for conformational flexibility. Additionally, Vina Dock employs a re-docking process to assess the best possible binding affinity of the generated molecules, while High Affinity measures the percentage of molecules that bind more strongly than the reference ligand for each test protein.
In terms of molecular properties, we assess drug-likeness using the Quantitative Estimation of Drug-likeness (QED) score\cite{bickerton2012quantifying}, where higher scores indicate better suitability for pharmaceutical applications. We evaluate synthetic accessibility (SA)\cite{ertl2009estimation}, with lower scores reflecting higher ease of synthesis. Finally, we measure chemical diversity by calculating the pairwise Tanimoto similarity of molecular fingerprints, ensuring the generated molecules are structurally diverse.
All evaluations are conducted on 100 novel protein-ligand complexes from the test set to ensure generalizability.

\subsection{Main Results}

\begin{table}[t]
\centering
\caption{Summary of various properties for reference molecules and those generated by our model and other baselines, where (↑) indicates that a larger value is better, and (↓) indicates that a smaller value is better. We highlight the best two results with \textbf{bold} text and \underline{underlined} text, respectively.}
\begin{tabular}{ccccccc}
\toprule
\hline
\\[-9pt]
\textbf{Model} & \textbf{QED} & \textbf{SA} & \textbf{Diversity} & \textbf{Vina Score} & \textbf{Vina Min} & \textbf{Vina Dock} \\ 
\hline 
\\[-9pt]
AR \cite{luo20213d} & \underline{0.51} & \underline{0.60} & 0.68 & \textbf{-5.75} & -6.18 & -6.75 \\ 
\hline
\\[-9pt]
GraphBP \cite{liu2022generating} & 0.43 & 0.49 & \textbf{0.79} & \textbackslash & \textbackslash & -4.80 \\ 
\hline
\\[-9pt]
TargetDiff \cite{guan20233d} & 0.48 & 0.58 & \underline{0.72} & -5.47 & -6.64 & -7.80 \\ 
\hline
\\[-9pt]
DecompDiff \cite{guan2024decompdiff} & 0.45 & 0.59 & 0.68 & \underline{-5.67} & \underline{-7.04} & \underline{-8.39} \\ 
\hline
\\[-9pt]
\textbf{Our Model} & \textbf{0.54} & \textbf{0.63} & 0.69 & -4.79 & \textbf{-7.07} & \textbf{-8.57} \\ 
\hline
\bottomrule
\end{tabular}
\end{table}

\begin{table}[t]
\centering
\caption{Jensen-Shannon divergence between the bond distance distributions of the reference molecules and the generated molecules, with lower values indicating better performance. The symbols '-', '=', and ':' represent single, double, and aromatic bonds, respectively. We highlight the best results with \textbf{bold} text.}
\begin{tabular}{cccccccc}
\toprule
\hline \\[-9pt]
\textbf{Model} & 6-6|4 & 6-6|1 & 6-8|1 & 6-7|1 & 6-8|2 & 6-6|2 & 6-7|4 \\ \hline \\[-9pt]
AR \cite{luo20213d} & 0.416 & 0.496 & 0.454 & 0.416 & 0.516 & 0.505 & 0.487 \\ \hline \\[-9pt]
GraphBP \cite{liu2022generating} & 0.407 & 0.368 & 0.467 & 0.456 & 0.471 & 0.530 & 0.689 \\ \hline \\[-9pt]
TargetDiff \cite{guan20233d} & 0.263 & 0.369 & 0.421 & 0.363 & 0.461 & 0.505 & 0.235 \\ \hline \\[-9pt]
DecompDiff \cite{guan2024decompdiff} & 0.251 & 0.359 & 0.376 & 0.344 & 0.374 & 0.537 & 0.269 \\ \hline \\[-9pt]
\textbf{Our Model} & \textbf{0.193} & \textbf{0.269} & \textbf{0.227} & \textbf{0.232} & \textbf{0.291} & \textbf{0.273} & \textbf{0.195} \\ 
\hline
\bottomrule
\end{tabular}
\end{table}
In this section, we present a comprehensive evaluation of our model across several key metrics, comparing it with existing state-of-the-art approaches: AR\cite{luo20213d}, GraphBP\cite{liu2022generating}, TargetDiff\cite{guan20233d}, and DecompDiff\cite{guan2024decompdiff}. These metrics include QED, SA, Diversity, as well as binding affinity measures such as Vina Score, Vina Min, Vina Dock, and the Jensen-Shannon distance for atom placement. Our model performs competitively across these metrics, showcasing its effectiveness in generating high-quality molecules with optimized binding affinity and desirable properties.

Our model achieves a mean QED score of 0.54, outperforming GraphBP\cite{liu2022generating}(0.43) and DecompDiff\cite{guan2024decompdiff}(0.45). 
This indicates that the generated molecules are not only drug-like but also exhibit higher alignment with the typical characteristics of pharmaceutical compounds. In terms of SA, our model's mean score of 0.63 ranks higher than GraphBP\cite{liu2022generating}(0.49) and TargetDiff\cite{guan20233d}(0.58), suggesting that the molecules are not only feasible to synthesize but also maintain good drug-like properties. Furthermore, our model demonstrates competitive Diversity with a mean score of 0.69, compared to TargetDiff\cite{guan20233d}(0.72) and GraphBP\cite{liu2022generating}(0.79), proving its ability to explore a broad chemical space while maintaining drug-likeness.

When comparing binding affinity, our model achieves the best performance in Vina Min and Vina Dock, indicating that the ligands generated by our method exhibit stronger binding affinity. For Jensen-Shannon distance, our model achieves the optimal results for all atomic distances, demonstrating that our method generates atom distributions that closely match the expected spatial geometry of the binding site, thus improving the relevance and accuracy of the generated molecules for structure-based drug design.

\subsection{Ablation Studies}
In this ablation study, we evaluate the impact of key components in our model. In the first experiment, we compare our full model, which uses a concept-based model to sample the arms, with a variant where the arms are directly taken from the reference molecule, and the molecule is then completed using a diffusion scaffold generation model. The results show that our concept-based approach significantly improves performance. Specifically, when the arms are sampled from the reference molecule, the performance metrics for QED, SA, Diversity, and Binding Affinity decrease, indicating that the concept-based sampling is crucial for generating high-affinity arms. These results highlight the importance of the concept-based model in guiding the generation of high-affinity arms, which is essential for optimizing drug-like properties.



\begin{table}[t]
\centering
\caption{Ablation study comparison between the full model and a variant using reference arms for QED, SA, and Diversity. We highlight the best results with bold text.}
\begin{tabular}{ccccccc}
\toprule
\hline \\[-9pt]
\textbf{Model} & \textbf{QED } & \textbf{SA } & \textbf{Diversity } & \textbf{Vina Score } & \textbf{Vina Min } & \textbf{Vina Dock }\\ \hline \\[-9pt]
Our Model & \textbf{0.54} & \textbf{0.63} & \textbf{0.69} & \textbf{-4.79} & \textbf{-7.07} & \textbf{-8.57}\\ \hline \\[-9pt]
Our Model ref\_arm & 0.48 & 0.60 & 0.37 & -3.94 & -5.39 & -6.89 \\ \hline
\bottomrule
\end{tabular}
\end{table}
\begin{table}[t]
\centering
\caption{Ablation study showing the importance of hydrogen bonds, hydrophobic contacts, and spatial information in Vina Dock performance. We highlight the best results with bold text.}
\begin{tabular}{ccccc}
\toprule
\hline \\[-9pt]
\textbf Model& Our Method & w.o. Hydrogen Bonds & w.o. Hydrophobic Contacts & w.o. Spatial Info \\ \hline \\[-9pt]
{Vina Dock} & \textbf{-8.57} & -5.80 & -6.96 & -7.01 \\ 
\hline
\bottomrule
\end{tabular}
\end{table}

In the second experiment, we evaluate the impact of various conceptual features by individually removing certain components from the model. These components include hydrogen bonds, hydrophobic contacts, and spatial information. We analyze how the removal of each of these concepts affects the overall performance of the model in terms of QED, SA, Diversity, and Binding Affinity. The results clearly show the importance of each concept, particularly the hydrogen bond and spatial information, which significantly contribute to both the drug-likeness (QED), synthetic accessibility and binding affinity (Vina scores). Removing these concepts leads to a noticeable decrease in performance, emphasizing their critical role in ensuring the generation of high-quality molecules.

\subsection{Qualitative analysis}
\begin{figure}[t]
    \centering
    \includegraphics[width=0.9\textwidth]{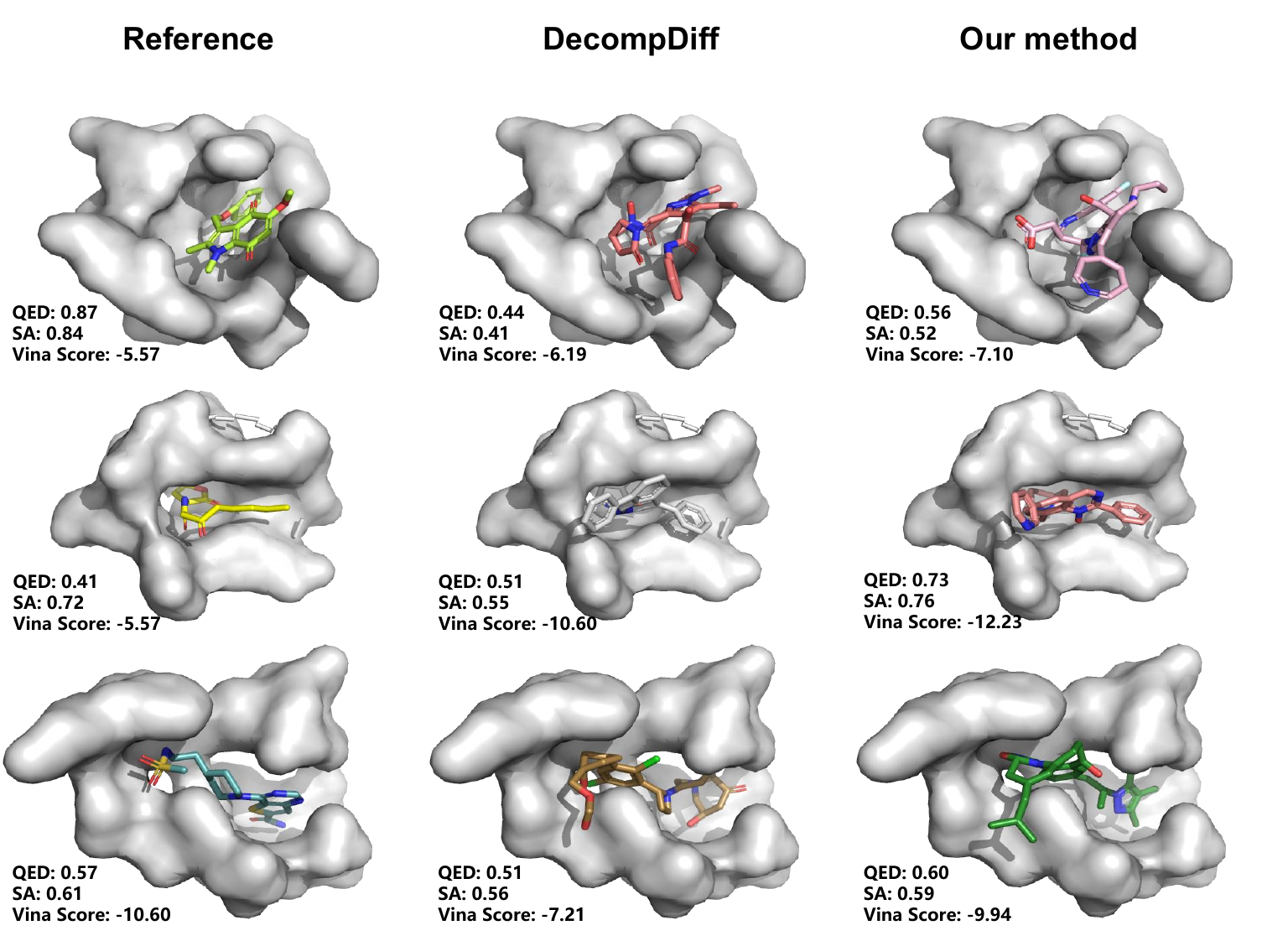}  
    \caption{\textbf{Comparison of molecular binding affinity across different methods.} This figure compares the molecular binding affinities of ligands generated by three different methods: the reference molecules, those produced by DecompDiff\cite{guan2024decompdiff}, and those generated by our proposed approach. Our method consistently outperforms DecompDiff\cite{guan2024decompdiff} in terms of binding affinity, drug-likeness (QED), and synthetic accessibility (SA). However, in cases where the protein sub-pocket is too small to accommodate a sufficient number of ligand arm atoms, the binding affinity tends to decrease, suggesting that our method’s performance could be further refined in such scenarios.}
    \label{fig:qa1}
\end{figure}
\begin{figure}[t]
    \centering
    \includegraphics[width=0.8\textwidth]{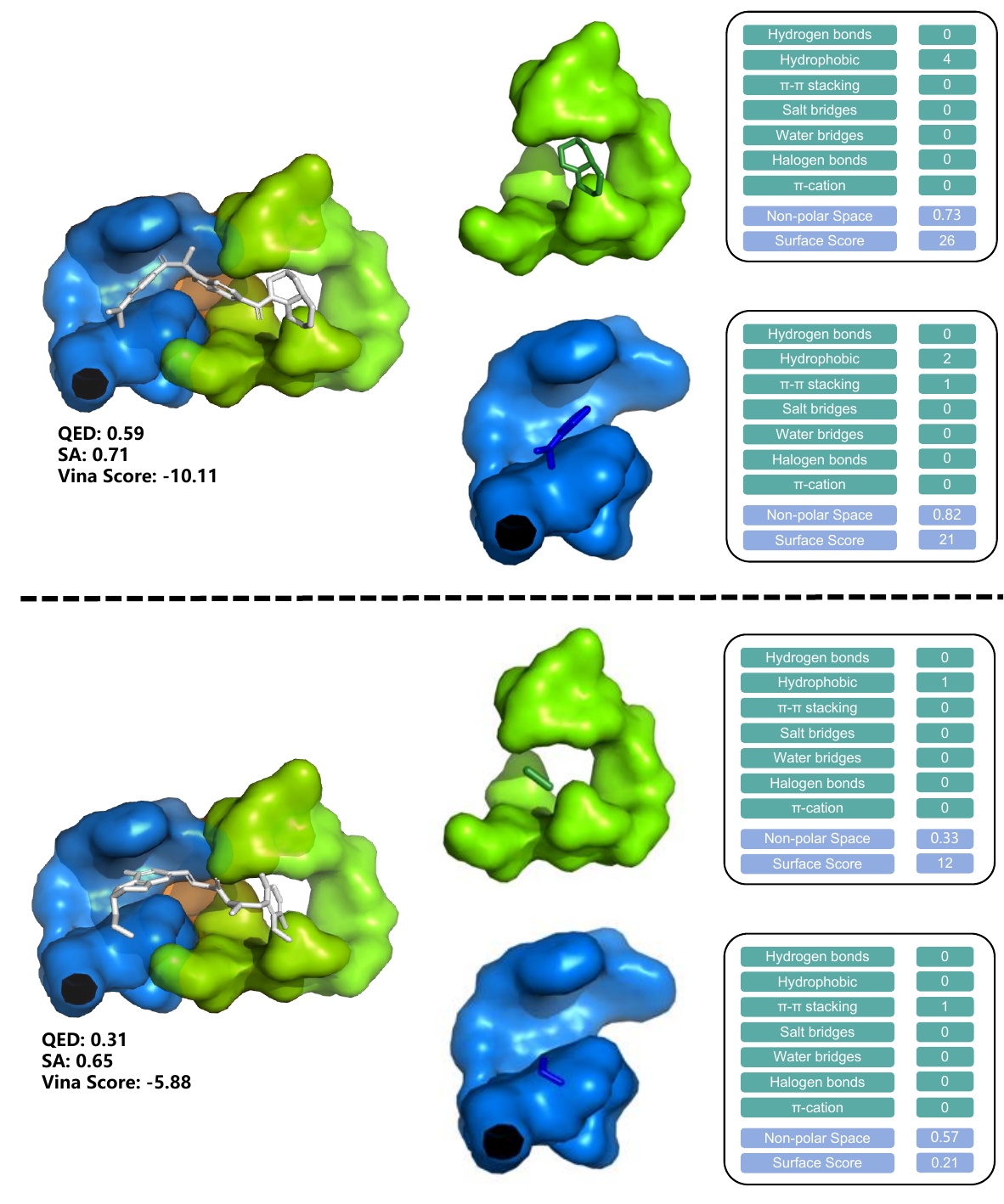}  
    \caption{\textbf{Comparison of molecular binding affinity across different arm from concepted-based model.} This figure illustrates the impact of ligand arm selection on molecular binding affinity. The top example shows optimal ligand arm choices, resulting in higher Vina scores, QED, and SA values, indicating strong binding affinity and synthetic feasibility. The bottom example demonstrates less optimal arm selection, leading to lower binding affinity and poorer drug-likeness, highlighting the importance of selecting high-quality ligand arms in our concept-based model.}
    \label{fig:qa2}
\end{figure}


In this section, we perform a qualitative analysis by comparing the ligand molecules generated by three methods: the reference ligand molecules, those generated by DecompDiff\cite{guan2024decompdiff}, and those generated by our method. Our approach not only generates molecules with superior binding affinity, as demonstrated by higher Vina scores, but also produces molecules that exhibit better synthetic feasibility and drug-likeness. The latter is indicated by the SA and QED metrics. These metrics suggest that the molecules produced by our method are not only more likely to bind effectively to their target proteins but are also easier to synthesize using current pharmaceutical methods, which is a critical factor in the practical application of these molecules in drug development.

As shown in the first and second examples in Figure \ref{fig:qa1}, our method consistently outperforms in both binding affinity (reflected by higher Vina scores), drug-likeness and synthetic accessibility. The superior binding affinity demonstrates that the ligand molecules generated by our model are more effective in interacting with their target protein sub-pockets, leading to stronger binding interactions. Furthermore, our method’s ability to produce molecules with a higher synthetic feasibility ensures that these molecules are more likely to be synthesized in a real-world setting using conventional pharmaceutical methods. This combination of strong binding potential and ease of synthesis makes our method particularly promising for drug design, where both effective binding and practical production are crucial factors for success.

However, despite the strengths of our approach, there are certain situations where it does not produce optimal ligand molecules. As shown in the third example in Figure \ref{fig:qa1}, the corresponding ligand arms formed based on the reference molecule’s protein sub-pocket are too small to accommodate a sufficient number of atoms in the ligand arms. In such cases, the sub-pocket's limited space restricts the possible interactions between the ligand arm and the protein, preventing the concept-based model from selecting the most suitable molecular arms. This results in a reduction of the ligand arm's influence on the protein sub-pocket, which, in turn, lowers the overall binding affinity of the generated ligand molecules. Thus, while our method performs well in many cases, it may not be suitable for all types of binding sites, particularly those with smaller or less adaptable sub-pockets. This limitation highlights an area for future improvement, where further refinement in handling such cases could make our method even more robust.

Additionally, as shown in Figure \ref{fig:qa2}, we further demonstrate the importance of the concepted-based model in selecting high-quality ligand arms. In the top example, the ligand arms selected using the model’s generated insights result in a well-fitting molecule, achieving superior Vina scores and favorable QED and SA metrics, indicating strong binding affinity and synthetic feasibility. In contrast, the bottom example, where less optimal ligand arms were chosen, results in a significantly lower Vina score and poorer drug-likeness. This comparison underscores the critical role of the concepted-based model in generating high-quality molecular arms that directly influence the binding affinity of the ligand molecules, reinforcing the idea that selecting the right molecular components is essential for designing potent and synthetically viable ligands.

\section{Conclusion}
We presented a novel 3D generative model for structure-based drug design that integrates a concept-based approach for arm sampling. Our method outperforms traditional models by generating higher-affinity molecules, demonstrating the importance of using concept-based sampling for optimizing drug-like properties. Ablation studies show that key features such as hydrogen bonds, salt bridges, hydrophobic contacts, and spatial information are critical for improving binding affinity and molecular quality. Our approach achieves superior performance in generating molecules with high binding affinity, favorable drug-likeness scores (QED), and synthetic accessibility, outperforming existing state-of-the-art models. 
However, our method’s performance diminishes when the protein pocket is too small to accommodate a sufficient number of ligand arm atoms, reducing binding affinity. Further improvements are needed to handle such cases and make the model more robust.

Future work will focus on scaling the model for larger protein pockets and incorporating multi-modal objectives to optimize both binding affinity and drug-like properties. We also plan to address the current limitation when the protein pocket can accommodate fewer ligand arm atoms. Developing strategies to improve the selection of suitable ligand arms for such pockets, or refining the model to account for these constraints, will be essential to enhancing the method's robustness and performance across a wider variety of protein targets.


\bibliographystyle{unsrt}  
\bibliography{references}

\end{document}